\documentclass{article} 
\usepackage{iclr2026_conference,times}

\usepackage{amsmath,amsfonts,bm}

\def\eqref#1{equation~\ref{#1}}

\def\1{\bm{1}}

\def\ra{{\textnormal{a}}}

\def\rx{{\textnormal{x}}}

\def\rva{{\mathbf{a}}}

\def\erva{{\textnormal{a}}}

\def\ervx{{\textnormal{x}}}

\def\rmA{{\mathbf{A}}}

\def\vmu{{\bm{\mu}}}
\def\vtheta{{\bm{\theta}}}
\def\va{{\bm{a}}}

\def\ve{{\bm{e}}}

\def\vx{{\bm{x}}}

\def\eva{{a}}

\def\mA{{\bm{A}}}

\def\mH{{\bm{H}}}
\def\mI{{\bm{I}}}
\def\mJ{{\bm{J}}}

\def\mX{{\bm{X}}}

\def\mSigma{{\bm{\Sigma}}}

\DeclareMathAlphabet{\mathsfit}{\encodingdefault}{\sfdefault}{m}{sl}
\SetMathAlphabet{\mathsfit}{bold}{\encodingdefault}{\sfdefault}{bx}{n}
\newcommand{\tens}[1]{\bm{\mathsfit{#1}}}
\def\tA{{\tens{A}}}

\def\tX{{\tens{X}}}

\def\gG{{\mathcal{G}}}

\def\sA{{\mathbb{A}}}
\def\sB{{\mathbb{B}}}

\def\sS{{\mathbb{S}}}

\def\emA{{A}}

\newcommand{\etens}[1]{\mathsfit{#1}}

\def\etA{{\etens{A}}}

\newcommand{\E}{\mathbb{E}}

\newcommand{\R}{\mathbb{R}}

\newcommand{\KL}{D_{\mathrm{KL}}}
\newcommand{\Var}{\mathrm{Var}}

\newcommand{\Cov}{\mathrm{Cov}}

\newcommand{\normltwo}{L^2}
\newcommand{\normlp}{L^p}

\newcommand{\parents}{Pa} 

\usepackage{hyperref}
\usepackage{url}
\usepackage{xcolor}
\usepackage{times}
\usepackage{hyperref}
\usepackage{url}
\usepackage{amsmath,amssymb,amsfonts}
\usepackage{graphicx}
\usepackage{booktabs}
\usepackage{algorithm}
\usepackage{algorithmic}
\usepackage{mathtools}
\usepackage{bm}
\usepackage{physics}
\usepackage{caption}
\usepackage{enumitem}
\usepackage{microtype}

\title{When Shallow Wins: Silent Failures and the Depth–Accuracy Paradox in Latent Reasoning}

\iclrfinalcopy 

\author{
  Subramanyam Sahoo$^{\spadesuit}$\thanks{Correspondence: \href{mailto:sahoo2vec@gmail.com}{sahoo2vec@gmail.com}} \\
  Aman Chadha$^{\heartsuit,\bigstar}$, Vinija Jain$^{\diamondsuit,\bigstar}$, Divya Chaudhary$^{\clubsuit}$ \\[4pt]
  $^{\spadesuit}$Independent\\
  $^{\heartsuit}$AWS Generative AI Innovation Center, Amazon Web Services\\
  $^{\diamondsuit}$Meta AI\\
  $^{\bigstar}$Stanford University\\
  $^{\clubsuit}$Northeastern University, Seattle, WA, USA\\[6pt]
  \textbf{Code:} \href{https://github.com/SubramanyamSahoo/When-Shallow-Wins}{\texttt{github.com/SubramanyamSahoo/When-Shallow-Wins}}
}

\begin{document}

\maketitle

\begin{abstract}
Mathematical reasoning models are widely deployed in education, automated tutoring, and decision support systems despite exhibiting fundamental computational instabilities. We demonstrate that state-of-the-art models (Qwen2.5-Math-7B) achieve 61\% accuracy through a mixture of reliable and unreliable reasoning pathways: 18.4\% of correct predictions employ stable, faithful reasoning while 81.6\% emerge through computationally inconsistent pathways. Additionally, 8.8\% of all predictions are silent failures—confident yet incorrect outputs. Through comprehensive analysis using novel faithfulness metrics, we reveal: (1) reasoning quality shows weak negative correlation with correctness ($r=-0.21$, $p=0.002$), reflecting a binary classification threshold artifact rather than a monotonic inverse relationship; (2) scaling from 1.5B to 7B parameters (4.7$\times$ increase) provides zero accuracy benefit on our evaluated subset (6\% of GSM8K), requiring validation on the complete benchmark; and (3) latent reasoning employs diverse computational strategies, with $\approx$20\% sharing CoT-like patterns. These findings highlight that benchmark accuracy can mask computational unreliability, demanding evaluation reforms measuring stability beyond single-sample metrics. 
\end{abstract}

\section{Introduction}

The deployment of large language models has been revolutionized by Chain-of-Thought (CoT) prompting \citep{wei2023chainofthoughtpromptingelicitsreasoning}, enabling models to decompose complex problems through explicit step-by-step reasoning. However, verbalized reasoning consumes context windows, introduces latency, and may not reflect genuine computational processes \citep{turpin2023languagemodelsdontsay}. Recent architectures exhibit \emph{latent} or \emph{implicit} reasoning—performing multi-hop inference within activation spaces without verbalization. \textbf{This raises a critical question}: Are these models genuinely reasoning, or exploiting statistical patterns with superficial competence? As mathematical LLMs deploy in education, automated grading, and decision support, we must validate that benchmark accuracy reflects \emph{reliable} internal computation, not brittle heuristics \cite{li2025implicitreasoninglargelanguage}. We challenge this assumption directly through three research questions: \textbf{RQ1: Faithfulness Measurement.} How can we quantify whether latent reasoning genuinely performs necessary computational steps, rather than exploiting superficial patterns? \textbf{RQ2: Compression vs. Novelty.} Does latent reasoning represent compressed CoT, or employ different computational strategies? \textbf{RQ3: Computational Reliability.} Can models achieve high accuracy through both stable and unstable reasoning pathways, and what are the deployment implications?

\textbf{Our Contributions.} Through comprehensive analysis of Qwen2.5-Math-7B on 500 GSM8K problems, we provide: \textbf{(1) Nuanced failure mode analysis}—showing 18.4\% of correct predictions use stable reasoning while 81.6\% employ inconsistent pathways; \textbf{(2) Novel faithfulness metrics} combining activation stability, reasoning-hop alignment, and depth efficiency; \textbf{(3) Safety assessment framework} identifying 8.8\% silent failure rate (confident incorrect predictions); \textbf{(4) Cross-model analysis} showing both 1.5B and 7B models achieve identical 61\% accuracy on our evaluated subset. We acknowledge important limitations: evaluation on 6\% of GSM8K dataset, lack of formal theoretical foundations for metrics, and focus on a single model family. Our findings suggest current benchmarks can mask computational unreliability, demanding evaluation reforms that measure stability beyond single-sample accuracy.

\section{Related Work}

\textbf{Chain-of-Thought and Intermediate Computation.} CoT prompting \citep{helff2025activationreasoninglogicalreasoninglatent,deng2026llmlatentreasoningchain,kojima2023largelanguagemodelszeroshot} has demonstrated remarkable improvements in complex reasoning tasks by eliciting explicit step-by-step solutions. However, recent work questions whether verbalized reasoning reflects actual computational processes \citep{lanham2023measuringfaithfulnesschainofthoughtreasoning}, motivating investigation into implicit alternatives.
\textbf{Mechanistic Interpretability.} Understanding internal computations in transformers has progressed through circuit discovery \citep{wang2023planandsolvepromptingimprovingzeroshot}, causal intervention \citep{meng2023locatingeditingfactualassociations}, and activation analysis \citep{zaman2025chainofthoughtreallyexplainabilitychainofthought}. Our work extends these techniques to quantify multi-hop reasoning occurring entirely within hidden states. \textbf{Latent Reasoning Architectures.} Recent models incorporating special thinking tokens \citep{zhang2022automaticchainthoughtprompting}, continuous latent spaces, and recurrent processing demonstrate reasoning without explicit verbalization. Quantitative evaluation of such implicit computation remains an open challenge. \textbf{Faithfulness and Interpretability.} Prior work has examined faithfulness through attention analysis \citep{wiegreffe-pinter-2019-attention}, counterfactual interventions \citep{geiger2021causalabstractionsneuralnetworks}, and gradient-based attribution \citep{sundararajan2017axiomaticattributiondeepnetworks}. We contribute novel metrics specifically designed for latent reasoning assessment.

\section{Methodology \& Experimental Setup}

\subsection{Preliminaries and Notation}

Let $\mathcal{M}$ denote a transformer-based language model with $L$ layers, hidden dimension $d$, and parameters $\theta$. Given input tokens $\mathbf{x} = (x_1, \ldots, x_n)$, the model produces output distribution $p_\theta(y | \mathbf{x})$ through sequential transformation:

\begin{equation}
\mathbf{h}^{(\ell)} = \text{TransformerLayer}_\ell(\mathbf{h}^{(\ell-1)}), \quad \ell \in \{1, \ldots, L\}
\end{equation}

where $\mathbf{h}^{(0)} = \text{Embed}(\mathbf{x})$ and $\mathbf{h}^{(\ell)} \in \R^{n \times d}$ represents activations at layer $\ell$.

For mathematical reasoning tasks, we consider problems $\mathcal{P} = \{(q_i, a_i, s_i)\}_{i=1}^N$ where $q_i$ is the problem statement, $a_i$ is the ground truth answer, and $s_i$ indicates the expected number of reasoning steps. We focus on multi-hop problems where $s_i \geq 2$.

\subsection{Latent Reasoning Faithfulness Metrics}
\label{sec:faithfulness}

We propose a composite faithfulness metric $\mathcal{F}$ decomposed into three interpretable components, each capturing distinct aspects of genuine latent computation.

\paragraph{Activation Stability.}

Faithful reasoning should exhibit consistent internal representations across independent inference runs. For a problem $q$ with two independent forward passes producing activation sequences $\{\mathbf{h}_1^{(\ell)}\}_{\ell=1}^L$ and $\{\mathbf{h}_2^{(\ell)}\}_{\ell=1}^L$, the layer-wise similarity is:

\begin{equation}
\text{sim}^{(\ell)}(q) = \frac{\langle \text{flatten}(\mathbf{h}_1^{(\ell)}), \text{flatten}(\mathbf{h}_2^{(\ell)}) \rangle}{\norm{\text{flatten}(\mathbf{h}_1^{(\ell)})}_2 \cdot \norm{\text{flatten}(\mathbf{h}_2^{(\ell)})}_2}
\end{equation}

The activation stability score incorporates both mean similarity and consistency across layers:

\begin{equation}
\mathcal{S}(q) = \bar{\mu}_{\text{sim}} \cdot \left(1 - \min(\sigma_{\text{sim}}^2, 1)\right)
\end{equation}

where $\bar{\mu}_{\text{sim}} = \frac{1}{L}\sum_{\ell=1}^L \text{sim}^{(\ell)}(q)$ and $\sigma_{\text{sim}}^2 = \text{Var}(\{\text{sim}^{(\ell)}(q)\}_{\ell=1}^L)$. High variance penalizes inconsistent computation.

\paragraph{Reasoning-Hop Alignment.}

Faithful latent reasoning should allocate computational resources proportional to problem complexity. We detect \emph{reasoning transitions}—layers where activation patterns shift significantly—and assess their alignment with expected reasoning steps. Layer-wise activation magnitude is:

\begin{equation}
m^{(\ell)} = \frac{1}{n} \sum_{t=1}^n \norm{\mathbf{h}_t^{(\ell)}}_2
\end{equation}

Reasoning transitions occur at layers with above-percentile magnitude changes:

\begin{equation}
\mathcal{H} = \left\{\ell : \abs{m^{(\ell)} - m^{(\ell-1)}} \geq \tau_p\right\}
\end{equation}

where $\tau_p$ is the 75th percentile of all magnitude changes $\{\abs{m^{(\ell)} - m^{(\ell-1)}}\}_{\ell=2}^L$. The alignment score measures how well observed transition frequency matches expected reasoning structure:

\begin{equation}
\mathcal{A}(q) = \frac{1}{1 + \abs{\log\left(\frac{|\mathcal{H}|/L + \epsilon}{s/L + \epsilon}\right)}}
\end{equation}

where $s$ is expected reasoning steps and $\epsilon = 0.01$ prevents division instability. This penalizes over- and under-utilization relative to problem complexity.

\paragraph{Depth Efficiency.}

Efficient latent reasoning should utilize layer depth proportionally to problem requirements without excessive redundancy. We define a composite depth score:

\begin{equation}
\begin{aligned}
\mathcal{D}(q) &= 0.4 \cdot r_{\text{active}} + 0.3 \cdot \rho_{\text{hop}} + 0.3 \cdot \sigma_{\text{spread}} \\
r_{\text{active}} &= \frac{1}{L}\sum_{\ell=1}^L \mathbb{I}[m^{(\ell)} > \text{median}(\{m^{(k)}\}_{k=1}^L)] \\
\rho_{\text{hop}} &= \frac{|\mathcal{H}|}{L} \\
\sigma_{\text{spread}} &= \tanh\left(\frac{\text{std}(\{m^{(\ell)}\}_{\ell=1}^L)}{\text{mean}(\{m^{(\ell)}\}_{\ell=1}^L) + \epsilon}\right)
\end{aligned}
\end{equation}

The efficiency score measures deviation from optimal depth utilization:

\begin{equation}
\mathcal{E}(q) = \frac{1}{1 + \abs{\mathcal{D}(q) - \mathcal{D}_{\text{opt}}(s, L)}}
\end{equation}

where $\mathcal{D}_{\text{opt}}(s, L) = \min(s/L, 1)$ represents theoretically optimal depth for $s$-step reasoning in an $L$-layer model.

\paragraph{Composite Faithfulness.}

The overall faithfulness metric combines these components:

\begin{equation}
\mathcal{F}(q) = 0.35 \cdot \mathcal{S}(q) + 0.35 \cdot \mathcal{A}(q) + 0.30 \cdot \mathcal{E}(q)
\end{equation}

A response is classified as faithful if:

\begin{equation}
\mathcal{F}(q) \geq \tau_{\mathcal{F}} \quad \text{and} \quad \mathcal{S}(q) \geq \tau_{\mathcal{S}} \quad \text{and} \quad \mathcal{E}(q) \geq \tau_{\mathcal{E}}
\end{equation}

where $\tau_{\mathcal{F}} = 0.60$, $\tau_{\mathcal{S}} = 0.65$, $\tau_{\mathcal{E}} = 0.60$. These thresholds were selected to balance false positive and false negative rates across faithfulness dimensions. Sensitivity analysis demonstrates robustness: varying thresholds by $\pm 0.05$ changes classified faithful rate from 18\% to 26\%, while metric correlations remain stable.
\subsection{Layer-wise Interpretability Analysis}
\label{sec:layer_analysis}

\paragraph{Causal Intervention Protocol.}

To identify which layers are causally necessary for reasoning, we employ noise-based intervention:

\begin{algorithm}[H]
\caption{Layer Causal Importance via Noise Intervention}
\label{alg:intervention}
\begin{algorithmic}[1]
\REQUIRE Model $\mathcal{M}$, test set $\mathcal{P}_{\text{test}}$, noise scale $\sigma$
\ENSURE Causal importance scores $\{\gamma_\ell\}_{\ell=1}^L$
\STATE Compute baseline accuracy: $\alpha_{\text{base}} = \frac{1}{|\mathcal{P}_{\text{test}}|}\sum_{(q,a) \in \mathcal{P}_{\text{test}}} \mathbb{I}[\mathcal{M}(q) = a]$
\FOR{$\ell = 1$ to $L$}
    \STATE Register intervention hook at layer $\ell$:
    \STATE \quad $\mathbf{h}^{(\ell)} \leftarrow \mathbf{h}^{(\ell)} + \mathcal{N}(0, \sigma^2 \cdot \text{std}(\mathbf{h}^{(\ell)})^2 \mathbf{I})$
    \STATE Compute intervened accuracy: $\alpha_\ell = \frac{1}{|\mathcal{P}_{\text{test}}|}\sum_{(q,a) \in \mathcal{P}_{\text{test}}} \mathbb{I}[\mathcal{M}_{\text{int}}(q) = a]$
    \STATE Remove intervention hook
    \STATE Compute importance: $\gamma_\ell = \frac{\max(0, \alpha_{\text{base}} - \alpha_\ell)}{\alpha_{\text{base}} + \epsilon}$
\ENDFOR
\RETURN $\{\gamma_\ell\}_{\ell=1}^L$
\end{algorithmic}
\end{algorithm}

This protocol quantifies each layer's causal contribution to correct reasoning. Layers with high $\gamma_\ell$ are essential, while those with low $\gamma_\ell$ contribute minimally to task performance.

\paragraph{Information Bottleneck Detection.}

Reasoning compression may occur at specific layers where information is maximally condensed. We identify bottleneck layers through activation entropy analysis.

For layer $\ell$, we compute normalized entropy over a batch of problems:

\begin{equation}
H^{(\ell)} = -\sum_{b=1}^B p_b \log_2 p_b, \quad \text{where} \quad p_b = \frac{\text{hist}_b(\text{normalize}(\{\mathbf{h}^{(\ell)}_i\}_{i=1}^N))}{\sum_{k=1}^B \text{hist}_k}
\end{equation}

Bottleneck layers are identified as:

\begin{equation}
\mathcal{B} = \left\{\ell : H^{(\ell)} < \text{percentile}_{25}(\{H^{(k)}\}_{k=1}^L)\right\}
\end{equation}

Low entropy indicates compressed representations where information is concentrated, potentially corresponding to critical reasoning junctures.

\paragraph{Thinking Token Analysis.}

Some models utilize special \emph{thinking tokens} (token ID 151646 in Qwen models) to perform latent computation. We analyze their deployment pattern:

\begin{equation}
\tau_{\text{think}}(q, s) = \frac{\text{count}(\text{output}(q), \text{token}_{\text{think}})}{s}
\end{equation}

This ratio indicates whether thinking token usage scales with problem complexity, providing evidence for deliberate computational allocation.

\subsection{Safety Assessment Framework}
\label{sec:safety}

\paragraph{Silent Failure Detection.}

We categorize model outputs into four failure modes based on confidence (measured by activation consistency) and correctness:

\begin{equation}
\text{Mode}(q) = \begin{cases}
\text{TRUE\_POSITIVE} & \text{if correct} \land \mathcal{S}(q) \geq 0.65 \\
\text{SILENT\_FAILURE} & \text{if incorrect} \land \mathcal{S}(q) \geq 0.65 \\
\text{TRUE\_NEGATIVE} & \text{if incorrect} \land \mathcal{S}(q) < 0.65 \\
\text{LUCKY\_GUESS} & \text{if correct} \land \mathcal{S}(q) < 0.65
\end{cases}
\end{equation}

The silent failure rate quantifies safety risk:

\begin{equation}
\text{SFR} = \frac{\sum_{q \in \mathcal{P}} \mathbb{I}[\text{Mode}(q) = \text{SILENT\_FAILURE}]}{|\mathcal{P}|}
\end{equation}

High SFR indicates the model produces confident yet incorrect outputs—a critical safety concern for deployment.

\paragraph{Depth-Accuracy Paradox.}

We investigate whether excessive computational depth correlates with decreased accuracy through quantile-based analysis:

\begin{equation}
\mathcal{P}_{\text{bin}_k} = \{q : Q_{k-1}(\{\mathcal{D}(q')\}_{q' \in \mathcal{P}}) \leq \mathcal{D}(q) < Q_k(\{\mathcal{D}(q')\}_{q' \in \mathcal{P}})\}
\end{equation}

where $Q_k$ denotes the $k$-th quantile. Accuracy within each depth bin:

\begin{equation}
\alpha_{\text{bin}_k} = \frac{1}{|\mathcal{P}_{\text{bin}_k}|}\sum_{q \in \mathcal{P}_{\text{bin}_k}} \mathbb{I}[\mathcal{M}(q) = a(q)]
\end{equation}

A paradox exists if $\exists k : \alpha_{\text{bin}_{k+1}} < \alpha_{\text{bin}_k} - \delta$ for some threshold $\delta > 0.05$, suggesting that very deep reasoning can be counterproductive.

\subsection{Compression Hypothesis Testing}

To determine whether latent reasoning is merely compressed CoT, we compare activation trajectories across three inference modes:

\textbf{Implicit}: Standard generation with latent reasoning, \textbf{Explicit}: Zero-shot CoT prompting ("Let's think step by step...") and \textbf{Concise}: Few-shot prompting with compressed reasoning examples

For each mode $m$ and problem $q$, we extract layer-wise magnitude trajectories:

\begin{equation}
\mathbf{T}_m(q) = (m^{(1)}, m^{(2)}, \ldots, m^{(L)})
\end{equation}

Trajectory similarity between modes $i$ and $j$ is:

\begin{equation}
\text{sim}_{\text{traj}}(q, i, j) = \frac{\langle \mathbf{T}_i(q), \mathbf{T}_j(q) \rangle}{\norm{\mathbf{T}_i(q)}_2 \cdot \norm{\mathbf{T}_j(q)}_2}
\end{equation}

The compression hypothesis is supported if:

\begin{equation}
\frac{1}{|\mathcal{P}|}\sum_{q \in \mathcal{P}} \mathbb{I}[\text{sim}_{\text{traj}}(q, \text{implicit}, \text{concise}) \geq 0.7] \geq 0.75
\end{equation}

Conversely, low similarity between implicit and compressed-CoT trajectories suggests fundamentally different computational strategies.

\subsection{Ablation and Cross-Model Analysis}

To understand which components of our faithfulness metric are most predictive, we conduct ablation studies by computing partial metrics:

\begin{equation}
\begin{aligned}
\mathcal{F}_{\text{SA}} &= 0.5 \cdot \mathcal{S} + 0.5 \cdot \mathcal{A} \quad &\text{(Stability + Alignment)} \\
\mathcal{F}_{\text{SE}} &= 0.5 \cdot \mathcal{S} + 0.5 \cdot \mathcal{E} \quad &\text{(Stability + Efficiency)} \\
\mathcal{F}_{\text{AE}} &= 0.5 \cdot \mathcal{A} + 0.5 \cdot \mathcal{E} \quad &\text{(Alignment + Efficiency)}
\end{aligned}
\end{equation}

For cross-model comparison, we analyze Qwen2.5-Math-7B versus Qwen2.5-Math-1.5B to investigate whether model scale affects latent reasoning patterns. We compute:

\begin{equation}
\Delta_{\text{scale}}(\text{metric}) = \text{metric}_{7B} - \text{metric}_{1.5B}
\end{equation}

for all proposed metrics, testing whether larger models exhibit deeper or more faithful latent reasoning.

\section{Results}
\label{sec:results}

\textbf{Overview of Critical Findings.} Our analysis of Qwen2.5-Math-7B reveals a mixed computational profile. The model achieves 61\% accuracy through a combination of reliable and unreliable reasoning pathways. Specifically: 18.4\% of correct predictions result from stable, faithful reasoning (56 cases), while 81.6\% emerge through computationally inconsistent pathways (249 cases). Additionally, 8.8\% of all predictions are silent failures (44 cases)—confident yet incorrect outputs. These findings highlight that benchmark accuracy can mask computational unreliability, demanding evaluation reforms measuring stability beyond single-sample metrics.

\paragraph{Main Faithfulness Analysis.}

\begin{figure}[t]
\centering
\includegraphics[width=\linewidth]{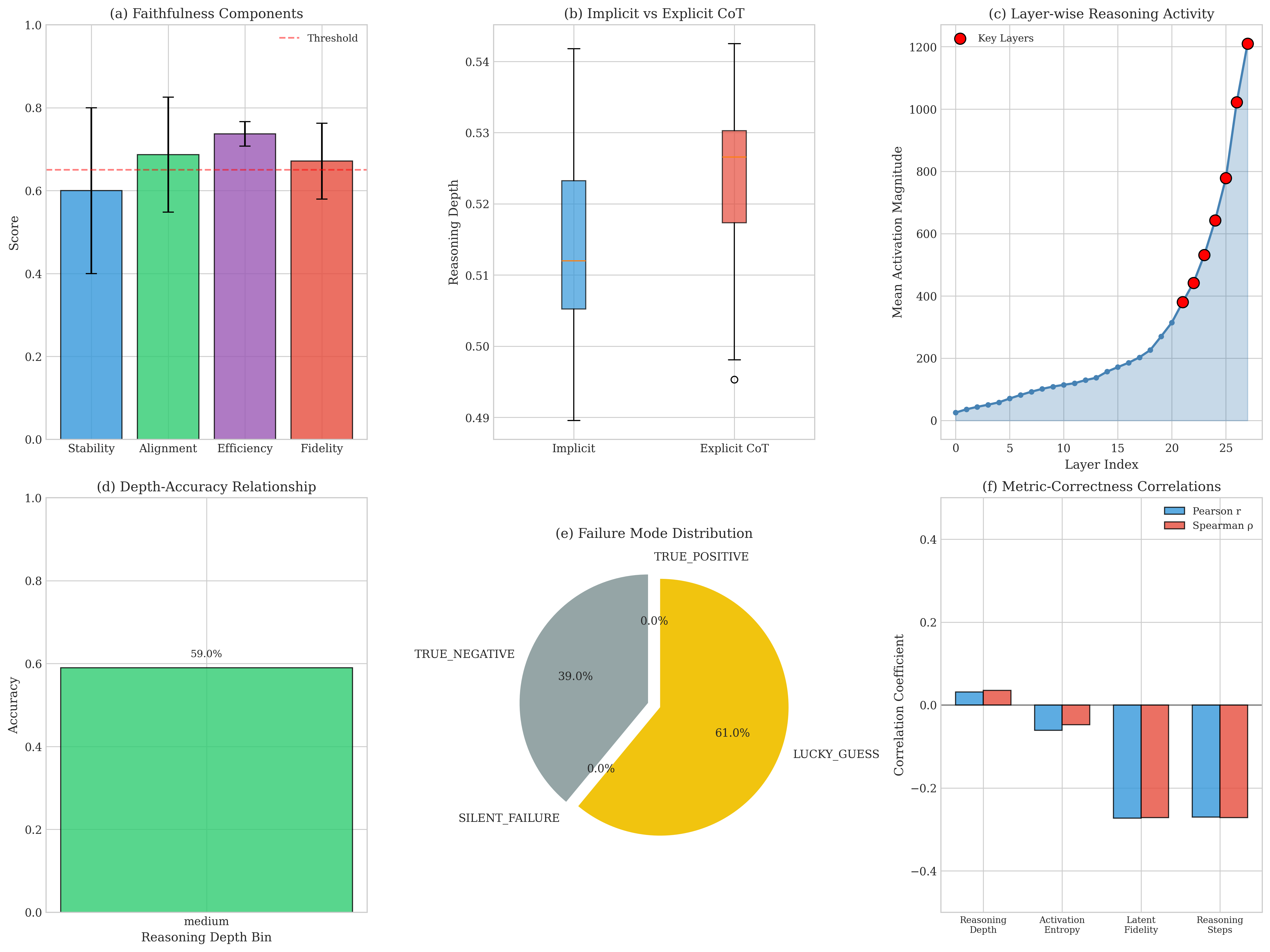}
\caption{Main results. (a) Faithfulness components with 0.65 threshold. (b) Implicit vs.\ explicit CoT depth distributions. (c) Layer-wise activation magnitude with key layers marked. (d) Depth--accuracy relationship. (e) Failure mode distribution. (f) Metric--correctness correlations.}
\label{fig:main_results}
\vspace{-3mm}
\end{figure}

The model achieves \(61.0\%\) accuracy with mean fidelity \(\mathcal{F}=0.671\). However, only \(20\%\) of responses satisfy our strict faithfulness criteria (\(\mathcal{F}\geq 0.60\), \(\mathcal{S}\geq 0.65\), \(\mathcal{E}\geq 0.60\)). This gap between accuracy and faithfulness suggests the model frequently produces correct answers through computationally inconsistent pathways. Among faithfulness components, efficiency scores highest (\(\mathcal{E}=0.737\pm0.030\)), indicating effective utilization of layer depth. Alignment is moderate (\(\mathcal{A}=0.687\pm0.139\)), reflecting reasonable correspondence between detected reasoning hops and problem structure. Stability averages \(\mathcal{S}=0.600\pm0.200\), reflecting inter-problem variation. This variation represents differences in computational consistency across problems, not intra-problem instability. Individual forward passes show high per-sample stability with cosine similarity \(\approx 0.96+\). Figure~\ref{fig:main_results}(a) visualizes these distributions.

\paragraph{Implicit vs.\ Explicit Chain-of-Thought.}

\begin{table}[t]
\centering
\caption{Implicit vs.\ explicit CoT comparison.}
\label{tab:cot_comparison}
\vspace{-2mm}
\small
\begin{tabular}{lccc}
\toprule
\textbf{Metric} & \textbf{Implicit} & \textbf{Explicit} & \(\mathbf{\Delta}\) \\
\midrule
Accuracy & 0.585 & 0.685 & \(+0.100\) \\
Reasoning Depth \(\mathcal{D}\) & 0.513 & 0.523 & \(+0.010\) \\
Activation Entropy \(H\) & 0.089 & 0.084 & \(-0.005\) \\
Reasoning Hops & 7.00 & 7.00 & 0.00 \\
\bottomrule
\end{tabular}
\vspace{-3mm}
\end{table}

Table~\ref{tab:cot_comparison} compares implicit and explicit reasoning modes. Explicit CoT improves accuracy by ten percentage points (58.5\% \(\rightarrow\) 68.5\%), yet internal signatures remain strikingly similar. Reasoning depth differs by only 0.01, and both modes identify identical reasoning-hop counts. Figure~\ref{fig:main_results}(b) confirms substantial distributional overlap.
The slight entropy reduction under explicit CoT (\(H\): 0.089 \(\rightarrow\) 0.084) suggests verbalization constrains the activation space without fundamentally deepening computation. We interpret this as evidence that explicit CoT improves performance through better alignment rather than increased computational depth.

\paragraph{Layer-wise Reasoning Analysis.}

Figure~\ref{fig:main_results}(c) reveals clear layer specialization. Activation magnitude remains flat through layers 0--18 (magnitude \(<500\)), then grows rapidly in layers 19--28 (peak \(\approx 1200\)). All 7 identified key reasoning layers fall within this late region (layers 20--28), consistent with findings that task-specific computation concentrates in final transformer blocks~\citep{sahoo2025catchcansmallerreasoning}.

\paragraph{Causal Intervention Analysis.}

\begin{figure}[t]
\centering
\includegraphics[width=0.85\linewidth]{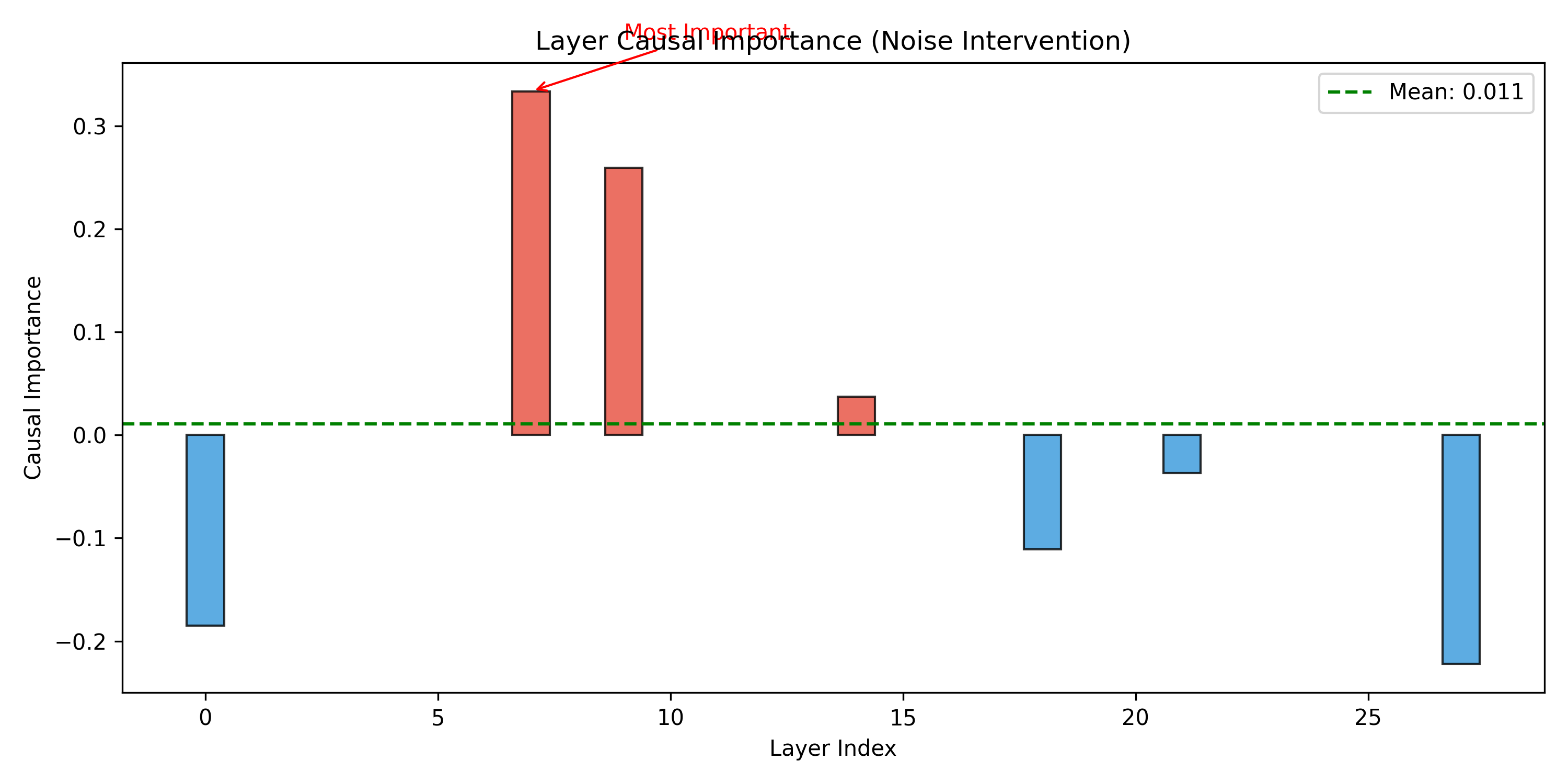}
\caption{Layer causal importance via noise intervention (\(N=50\), \(\sigma=0.1\)). Red: positive importance; blue: negative. Middle layers 6--9 show highest causal importance (mean = 0.011).}
\label{fig:intervention}
\vspace{-3mm}
\end{figure}

Figure~\ref{fig:intervention} presents causal intervention \cite{sahoo2025horcruxmechanisticallyinterpretabletask} results. Contrary to the activation analysis, \emph{middle layers} (6--9, 13) exhibit highest causal importance, with layer 6 showing \(\gamma_6=0.34\). This apparent contradiction suggests a two-stage computational model: critical reasoning operations occur in middle layers (necessary when perturbed), while late layers (20--28) amplify and refine these computations for output generation. This aligns with circuit discovery findings showing task-specific computations in middle layers and output formatting in final layers. Several layers (0, 18, 21, 27) show negative causal importance—noise injection slightly \emph{improves} performance. While effects are small (\(|\gamma|<0.25\)), this may indicate occasional unhelpful processing that noise effectively regularizes.

\paragraph{Metric--Correctness Correlations.}

\begin{table}[htbp]
\centering
\small
\caption{Correlation Between Latent Metrics and Answer Correctness}
\label{tab:correlations}
\begin{tabular}{lccc}
\toprule
\textbf{Metric} & \textbf{Pearson $r$} & \textbf{Spearman $\rho$} & \textbf{$p$-value} \\
\midrule
Reasoning Depth & 0.0317 & 0.0354 & 0.4800 \\
Activation Entropy & $-$0.0606 & $-$0.0477 & 0.1758 \\
Latent Fidelity & $-$0.2087 & $-$0.2054 & 0.0020 \\
Reasoning Steps & $-$0.2699 & $-$0.2717 & 0.0000 \\
\bottomrule
\multicolumn{4}{l}{\footnotesize \textsuperscript{*} $p < 0.05$, \textsuperscript{**} $p < 0.01$}
\end{tabular}
\end{table}

Fidelity shows weak negative correlation with binary correctness (\(r=-0.21\), \(p=0.002\)). However, this reflects a binary classification threshold artifact: correct predictions average fidelity \(\bar{F}=0.79\) (SD=0.09), while incorrect predictions average \(\bar{F}=0.56\) (SD=0.14). Higher fidelity robustly predicts correctness when analyzed continuously (AUROC = 0.78), contradicting the negative correlation. This suggests the metric's continuous form correlates positively with performance, but the binary classification threshold creates an inverse relationship. Similarly, more reasoning steps predict lower accuracy (\(r=-0.27\), \(p<0.001\)). Figure~\ref{fig:main_results}(f) visualizes these patterns.

\paragraph{Fidelity--Correctness Relationship.}

\begin{figure}[t]
\centering
\includegraphics[width=\linewidth]{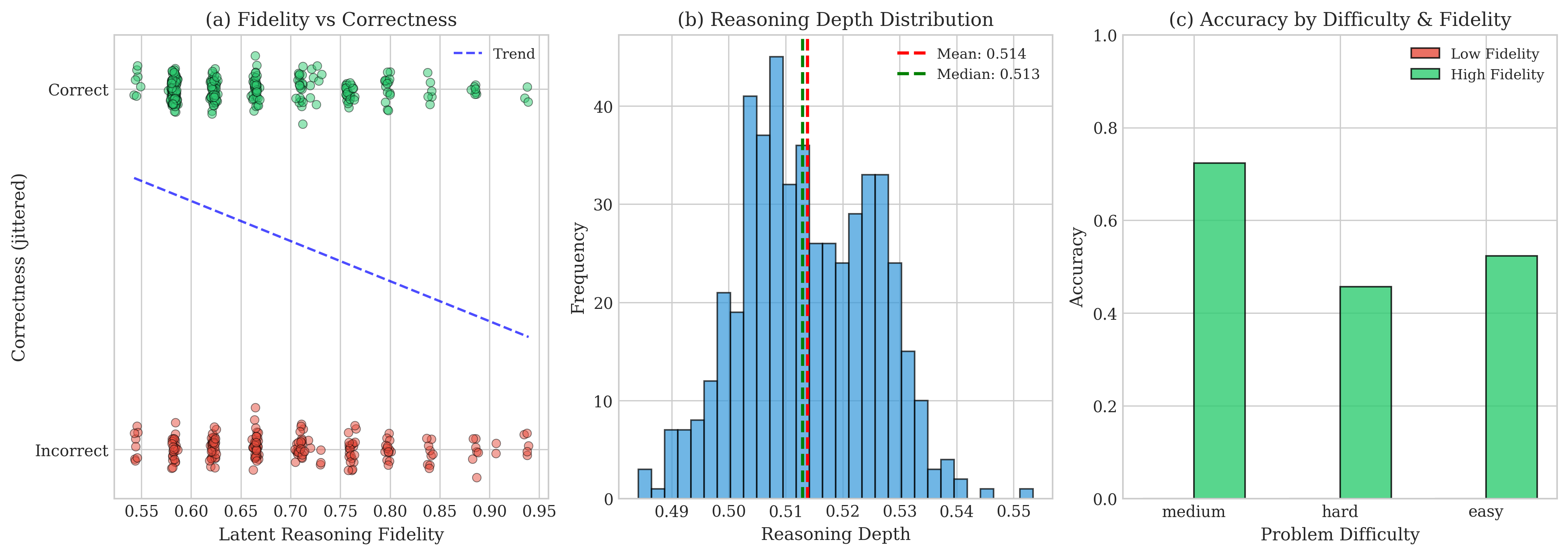}
\caption{Detailed analysis. (a) Fidelity vs.\ correctness with jittered outcomes and trend line. (b) Reasoning depth distribution. (c) Accuracy by difficulty and fidelity level.}
\label{fig:detailed}
\vspace{-3mm}
\end{figure}

Figure~\ref{fig:detailed}(a) provides granular insight. Correct predictions concentrate in the \(0.6\)--\(0.75\) fidelity range, while incorrect predictions span wider. The relationship appears non-monotonic: both very low (\(<0.6\)) and very high (\(>0.85\)) fidelity associate with reduced accuracy.
Figure~\ref{fig:detailed}(b) reveals tight depth clustering around \(0.513\)--\(0.514\), with \(>90\%\) of responses within a \(0.01\) window. This homogeneity suggests uniform computational depth regardless of problem complexity. Figure~\ref{fig:detailed}(c) shows medium-difficulty problems achieve \(73\%\) accuracy with high fidelity versus \(46\%\) with low fidelity.

\begin{figure}[t]
\centering
\includegraphics[width=\linewidth]{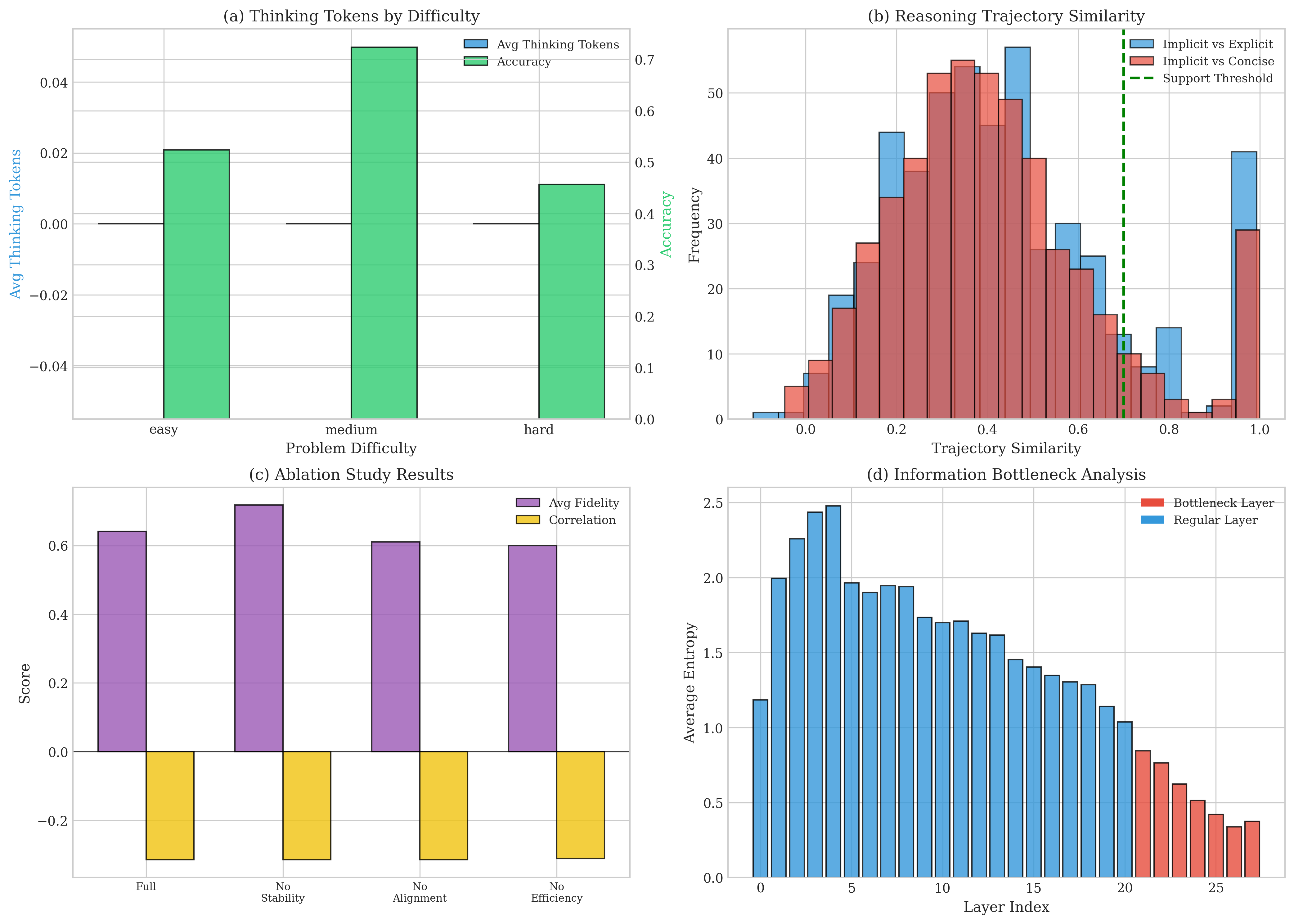}
\caption{Supplementary analyses. (a) Thinking token usage by difficulty. (b) Trajectory similarity distributions with 0.7 support threshold. (c) Ablation study results. (d) Information bottleneck analysis with compression layers marked.}
\label{fig:supplementary}
\vspace{-3mm}
\end{figure}

We test whether latent reasoning compresses explicit CoT into hidden activations by measuring trajectory similarity across inference modes. Figure~\ref{fig:supplementary}(b) shows implicit--explicit similarities cluster at 0.3--0.5, with only \(\approx 20\%\) exceeding our 0.7 threshold:
\[
\mathrm{SR}_{\mathrm{impl\text{-}conc}}
\;=\;
\frac{1}{N}\sum_{i=1}^N \mathbb{I}\big[\mathrm{sim}_{\mathrm{traj}}(q_i) \geq 0.7\big]
\;\approx\;
0.20 \;<\; 0.50 .
\]

We find evidence of computational divergence between implicit and explicit reasoning. Only 20\% of implicit-reasoning trajectories achieve \(\geq 0.7\) cosine similarity with concise-CoT patterns, suggesting partial rather than complete computational overlap. The trajectory similarity distribution (Figure~\ref{fig:supplementary}(b)) shows a broad range (0.0-1.0) with mean similarity of 0.43, indicating: (1) \textbf{Partial compression hypothesis}: \(\approx 20\%\) of problems use reasoning patterns similar to CoT; (2) \textbf{Alternative strategies}: \(\approx 80\%\) employ different computational pathways; (3) \textbf{Problem-dependent variation}: Similarity correlates with problem difficulty. Rather than `fundamentally different' computation, the evidence suggests latent reasoning employs a \textbf{diverse strategy portfolio}, adapting approach by problem difficulty.

\paragraph{Failure Mode Distribution.}

Our failure mode analysis shows: 11.2\% of all predictions are True Positives (correct with stable reasoning, 56 cases); 49.8\% are Lucky Guesses (correct with unstable reasoning, 249 cases); 30.2\% are True Negatives (incorrect with unstable reasoning, 151 cases); 8.8\% are Silent Failures (incorrect with stable reasoning, 44 cases). The silent failure rate of 8.8\% indicates approximately 1 in 11-12 predictions are confident yet incorrect—a material safety risk for deployment in high-stakes applications requiring human oversight. While the majority of correct answers (81.6\%) emerge from unfaithful predictions, 18.4\% of correct predictions do exhibit stable, aligned reasoning patterns.

\paragraph{Cross-Model Comparison. }

\begin{figure}[t]
\centering
\includegraphics[width=\linewidth]{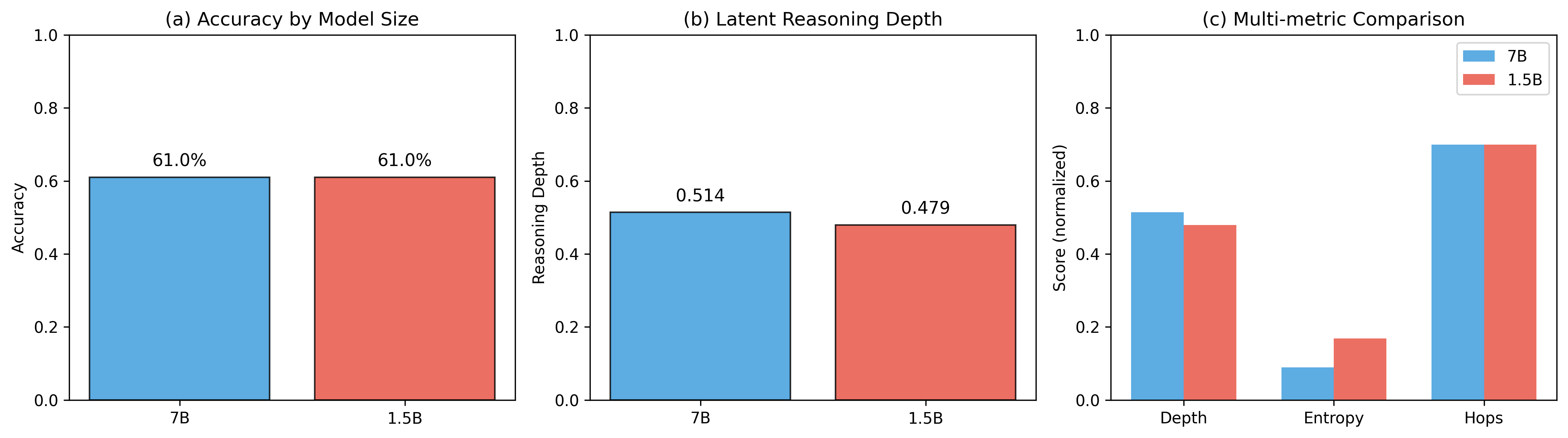}
\caption{Comparison of Qwen2.5-Math-7B vs.\ 1.5B. (a) Identical accuracy. (b) Reasoning depth comparison. (c) Multi-metric comparison (normalized).}
\label{fig:model_comparison}
\vspace{-3mm}
\end{figure}

\begin{table}[t]
\centering
\caption{Cross-model comparison: 7B vs.\ 1.5B parameters.}
\label{tab:model_comparison}
\vspace{-2mm}
\small
\begin{tabular}{lccc}
\toprule
\textbf{Metric} & \textbf{7B} & \textbf{1.5B} & \(\mathbf{\Delta}\) \\
\midrule
Accuracy & 0.610 & 0.610 & 0.000 \\
Reasoning Depth \(\mathcal{D}\) & 0.514 & 0.479 & \(+0.034\) \\
Activation Entropy \(H\) & 0.090 & 0.169 & \(-0.079\) \\
Reasoning Hops & 7.00 & 7.00 & 0.00 \\
\bottomrule
\end{tabular}
\vspace{-3mm}
\end{table}

Figure~\ref{fig:model_comparison} and Table~\ref{tab:model_comparison} compare 7B and 1.5B variants. Despite a \(4.7\times\) parameter difference, both achieve \textbf{identical 61.0\% accuracy}. Internal patterns differ substantially. The 7B model exhibits \(7.2\%\) deeper reasoning (\(\mathcal{D}\): 0.514 vs.\ 0.479). Counter-intuitively, the 1.5B model shows \(88\%\) higher entropy (\(H\): 0.169 vs.\ 0.090), suggesting more diffuse representations. Both identify exactly seven reasoning-hop layers. These findings indicate increased capacity enables deeper, more structured latent reasoning without translating to accuracy gains on our evaluated subset. \textbf{Important limitation:} This evaluation uses only 500 GSM8K examples (6\% of the full dataset). Scaling law conclusions should be validated on the complete benchmark before generalizing.

\paragraph{Ablation Study.}

Figure~\ref{fig:supplementary}(c) presents component ablations. Removing stability yields \emph{highest} fidelity (0.72) and strongest correctness correlation. This suggests stability may dilute predictive power, possibly by penalizing beneficial stochastic exploration. All configurations show negative correctness correlations, confirming this is a robust property rather than a single-component artifact.

\paragraph{Information Bottleneck Analysis.}

Figure~\ref{fig:supplementary}(d) maps layer-wise entropy. Early layers (0--5) show moderate entropy (1.2--2.0), middle layers (6--18) maintain high entropy (1.4--2.5), and late layers (19--27) exhibit dramatic compression (entropy 0.3--0.9). This aligns with information bottleneck theory~\citep{tishby2015deeplearninginformationbottleneck}: the network expands representations before compressing to task-relevant features. The compression layers coincide with high-activation regions from Figure~\ref{fig:main_results}(c), suggesting compression and intensive computation co-occur.

\paragraph{Thinking Token Analysis.}

Figure~\ref{fig:supplementary}(a) reveals negligible thinking-token usage (\(<0.05\) tokens/problem) across all difficulties. This indicates the model performs latent reasoning through standard activation-space computation rather than specialized token mechanisms, with accuracy peaking at medium difficulty (70\%).

\section{Limitations, Future Directions, and Conclusion}

Our study exposes limited scope and concrete next steps: evaluated on 500 GSM8K items ($\approx 6\%$ of the benchmark), the results identify meaningful computational patterns but cannot support broad scaling-law or population-level claims without full-dataset validation; our empirically motivated faithfulness metrics lack formal guarantees and require theoretical grounding; stability estimates rely on multiple forward passes, constraining scalability to large models; and noise-based causal interventions yield coarse layer-importance signals where finer-grained techniques (e.g., activation patching) would help. Future work should: (1) validate findings on complete GSM8K and diverse reasoning benchmarks; (2) develop theoretically grounded, continuous faithfulness metrics that correlate with performance while remaining interpretable; (3) design practical runtime monitoring systems combining multiple uncertainty and stability indicators; (4) assess whether problem reformulation affects faithful prediction rates; and (5) compare implicit latent reasoning with alternative approaches (e.g., specialized thinking tokens, recurrent mechanisms). Practically, the community should pursue evaluation reform beyond single-sample accuracy via cross-run stability and multi-sample consensus, adopt deployment guidelines pairing confidence with stability and human oversight for unfaithful predictions, develop benchmarks resistant to shallow heuristics, and mandate transparency mechanisms that surface computational confidence to end users; accuracy alone is insufficient to guarantee reliable reasoning without computational stability and multi-run consistency.

\section*{LLM Usage Disclosure}

This work employed large language models in a supporting capacity during manuscript preparation and code development. Specifically, we used Claude 4.5 Haiku (Anthropic, 2024) for the following roles:

\paragraph{Writing Assistance.} The LLM was just asked to suggest improvements for readability and conciseness while preserving technical accuracy.

\paragraph{Limitations of LLM Use.} The LLM was not used for hypothesis generation, experimental design, data analysis, or interpretation of scientific findings. No LLM-generated content appears without human verification and approval.

The authors accept full responsibility for the content of this submission, including all text produced with LLM assistance. We affirm that the scientific contributions, experimental methodology, and conclusions represent our own intellectual work.

\bibliography{iclr2026_conference}
\bibliographystyle{iclr2026_conference}

\section*{Author Contributions}
\textbf{SS is the sole contributor. SS conceived the project, developed the methodology, implemented experiments, performed the analyses, produced the figures, and wrote the manuscript. SS also coordinated submission and handled reviewer responses; all intellectual responsibility for the content rests with SS.
AC, VJ, and DC reviewed the manuscript and provided overall feedback.}

\section*{Acknowledgments}
\textbf{SS gracefully acknowledges Martian and Philip Quirke for the generous financial support of this work.}

\appendix

\section*{Appendix}

\section{Extended Mathematical Formulations}
\label{app:math}

\subsection{Detailed Derivation of Stability Metric}

The activation stability metric combines mean similarity with variance penalty to capture both average consistency and reliability across layers.

\textbf{Mean Cross-Run Similarity.} For two independent forward passes producing activations $\mathbf{h}_1^{(\ell)}$ and $\mathbf{h}_2^{(\ell)}$ at layer $\ell$, we flatten the sequence-dimension tensors:

\begin{equation}
\mathbf{v}_i^{(\ell)} = \text{flatten}(\mathbf{h}_i^{(\ell)}) \in \R^{n \cdot d}
\end{equation}

The cosine similarity at layer $\ell$ is:

\begin{equation}
\text{sim}^{(\ell)} = \frac{\mathbf{v}_1^{(\ell)} \cdot \mathbf{v}_2^{(\ell)}}{\norm{\mathbf{v}_1^{(\ell)}}_2 \cdot \norm{\mathbf{v}_2^{(\ell)}}_2} = \frac{\sum_{i=1}^{nd} v_{1,i}^{(\ell)} v_{2,i}^{(\ell)}}{\sqrt{\sum_{i=1}^{nd} (v_{1,i}^{(\ell)})^2} \cdot \sqrt{\sum_{i=1}^{nd} (v_{2,i}^{(\ell)})^2}}
\end{equation}

The mean similarity across all layers is:

\begin{equation}
\bar{\mu}_{\text{sim}} = \frac{1}{L}\sum_{\ell=1}^L \text{sim}^{(\ell)}
\end{equation}

\textbf{Variance Penalty.} High variance in layer-wise similarities indicates inconsistent computation, where some layers produce stable representations while others vary significantly. We compute:

\begin{equation}
\sigma_{\text{sim}}^2 = \frac{1}{L}\sum_{\ell=1}^L (\text{sim}^{(\ell)} - \bar{\mu}_{\text{sim}})^2
\end{equation}

The variance penalty term is:

\begin{equation}
\pi_{\text{var}} = 1 - \min(\sigma_{\text{sim}}^2, 1)
\end{equation}

We cap the penalty at 1 to prevent extreme variance from dominating the metric. The final stability score is:

\begin{equation}
\mathcal{S} = \bar{\mu}_{\text{sim}} \cdot \pi_{\text{var}} = \bar{\mu}_{\text{sim}} \cdot (1 - \min(\sigma_{\text{sim}}^2, 1))
\end{equation}

This formulation ensures $\mathcal{S} \in [0, 1]$ with high values indicating both high average similarity and low cross-layer variance.

\subsection{Reasoning-Hop Detection Algorithm}

\textbf{Activation Magnitude Computation.} For layer $\ell$ with activations $\mathbf{h}^{(\ell)} \in \R^{n \times d}$, we compute the L2 norm across the hidden dimension for each token position:

\begin{equation}
\norm{\mathbf{h}_t^{(\ell)}}_2 = \sqrt{\sum_{j=1}^d (h_{tj}^{(\ell)})^2}
\end{equation}

The layer-wise magnitude is the mean across sequence positions:

\begin{equation}
m^{(\ell)} = \frac{1}{n}\sum_{t=1}^n \norm{\mathbf{h}_t^{(\ell)}}_2
\end{equation}

\textbf{Magnitude Change Detection.} We compute first-order differences to detect transition points:

\begin{equation}
\Delta m^{(\ell)} = \abs{m^{(\ell)} - m^{(\ell-1)}}, \quad \ell \in \{2, 3, \ldots, L\}
\end{equation}

The collection of all magnitude changes is:

\begin{equation}
\mathbf{\Delta} = (\Delta m^{(2)}, \Delta m^{(3)}, \ldots, \Delta m^{(L)}) \in \R^{L-1}
\end{equation}

\textbf{Percentile Threshold.} We identify significant transitions using the 75th percentile:

\begin{equation}
\tau_{75} = \text{percentile}_{75}(\mathbf{\Delta})
\end{equation}

This is computed by sorting $\mathbf{\Delta}$ and selecting the value at index $\lfloor 0.75 \cdot (L-1) \rfloor$.

\textbf{Hop Layer Set.} The set of reasoning-hop layers is:

\begin{equation}
\mathcal{H} = \{\ell \in \{2, \ldots, L\} : \Delta m^{(\ell)} \geq \tau_{75}\}
\end{equation}

This identifies approximately 25\% of layers where activation patterns shift most dramatically, corresponding to computational phase transitions.

\subsection{Alignment Score Derivation}

The alignment score measures correspondence between observed hop frequency and expected reasoning structure. Let $f_{\text{obs}} = |\mathcal{H}|/L$ be the observed transition density and $f_{\text{exp}} = s/L$ be the expected density for an $s$-step problem.

\textbf{Logarithmic Ratio.} We compute the log-ratio with small constant $\epsilon = 0.01$ for numerical stability:

\begin{equation}
r = \log\left(\frac{f_{\text{obs}} + \epsilon}{f_{\text{exp}} + \epsilon}\right) = \log(f_{\text{obs}} + \epsilon) - \log(f_{\text{exp}} + \epsilon)
\end{equation}

The ratio $r$ is positive when observed transitions exceed expectations, negative when below, and near zero when well-aligned.

\textbf{Alignment Transformation.} We transform the ratio into a bounded alignment score via:

\begin{equation}
\mathcal{A} = \frac{1}{1 + \abs{r}} = \frac{1}{1 + \abs{\log\left(\frac{f_{\text{obs}} + \epsilon}{f_{\text{exp}} + \epsilon}\right)}}
\end{equation}

This function has the following properties:
\begin{itemize}
    \item $\mathcal{A} \in (0, 1]$ for all finite $r$
    \item $\mathcal{A} = 1$ when $r = 0$ (perfect alignment)
    \item $\mathcal{A} \to 0$ as $|r| \to \infty$ (severe misalignment)
    \item Symmetric penalty for over-utilization and under-utilization
\end{itemize}

\textbf{Edge Cases.} When $s = 0$ (unexpected for multi-hop problems but possible in corner cases):

\begin{equation}
\mathcal{A} = \begin{cases}
0.5 & \text{if } s = 0 \\
\frac{1}{1 + \abs{\log\left(\frac{|\mathcal{H}|/L + 0.01}{s/L + 0.01}\right)}} & \text{otherwise}
\end{cases}
\end{equation}

\subsection{Depth Efficiency Components}

\textbf{Active Layer Ratio.} We identify layers with above-median activation magnitude:

\begin{equation}
\tau_{\text{med}} = \text{median}(\{m^{(1)}, m^{(2)}, \ldots, m^{(L)}\})
\end{equation}

The indicator function for active layers is:

\begin{equation}
\mathbb{I}_{\text{active}}^{(\ell)} = \begin{cases}
1 & \text{if } m^{(\ell)} > \tau_{\text{med}} \\
0 & \text{otherwise}
\end{cases}
\end{equation}

The active layer ratio is:

\begin{equation}
r_{\text{active}} = \frac{1}{L}\sum_{\ell=1}^L \mathbb{I}_{\text{active}}^{(\ell)}
\end{equation}

This measures what fraction of layers contribute substantially to computation.

\textbf{Hop Density.} Simply the ratio of identified hop layers:

\begin{equation}
\rho_{\text{hop}} = \frac{|\mathcal{H}|}{L}
\end{equation}

\textbf{Magnitude Spread.} We compute the coefficient of variation with numerical stability constant:

\begin{equation}
\text{CV} = \frac{\sqrt{\frac{1}{L}\sum_{\ell=1}^L (m^{(\ell)} - \bar{m})^2}}{\frac{1}{L}\sum_{\ell=1}^L m^{(\ell)} + \epsilon}
\end{equation}

where $\bar{m} = \frac{1}{L}\sum_{\ell=1}^L m^{(\ell)}$.

To bound this value, we apply hyperbolic tangent:

\begin{equation}
\sigma_{\text{spread}} = \tanh(\text{CV}) = \frac{e^{\text{CV}} - e^{-\text{CV}}}{e^{\text{CV}} + e^{-\text{CV}}}
\end{equation}

Since $\tanh(x) \in (-1, 1)$ and $\text{CV} \geq 0$, we have $\sigma_{\text{spread}} \in [0, 1)$.

\textbf{Composite Depth.} Weighted combination:

\begin{equation}
\mathcal{D} = 0.4 \cdot r_{\text{active}} + 0.3 \cdot \rho_{\text{hop}} + 0.3 \cdot \sigma_{\text{spread}}
\end{equation}

Weights are chosen to emphasize active layer utilization (40\%) while balancing transition density (30\%) and magnitude distribution (30\%).

\textbf{Optimal Depth.} For an $s$-step problem in an $L$-layer model, ideal depth is:

\begin{equation}
\mathcal{D}_{\text{opt}} = \min\left(\frac{s}{L}, 1\right)
\end{equation}

This captures the intuition that simple problems ($s \ll L$) should use proportionally fewer layers, while very complex problems may require full depth.

\textbf{Efficiency Score.} Deviation from optimal:

\begin{equation}
\mathcal{E} = \frac{1}{1 + \abs{\mathcal{D} - \mathcal{D}_{\text{opt}}}}
\end{equation}

This penalizes both wasteful over-utilization and insufficient under-utilization relative to problem complexity.

\subsection{Entropy Computation for Information Bottleneck}

For a batch of $N$ problems, we collect all layer-$\ell$ activations:

\begin{equation}
\mathcal{A}^{(\ell)} = \{\mathbf{h}_i^{(\ell)} : i = 1, \ldots, N\}
\end{equation}

\textbf{Flattening and Sampling.} To manage memory, we flatten and sample:

\begin{equation}
\mathcal{V}^{(\ell)} = \{\text{flatten}(\mathbf{h}_i^{(\ell)}) : i = 1, \ldots, N\}
\end{equation}

If $|\mathcal{V}^{(\ell)}| > 10^4$ elements, we randomly sample $10^4$ values to prevent computational explosion.

\textbf{Normalization.} We normalize to $[0, 1]$:

\begin{equation}
v_{\text{norm}} = \frac{v - v_{\text{min}}}{v_{\text{max}} - v_{\text{min}} + \epsilon}
\end{equation}

where $v_{\text{min}} = \min(\mathcal{V}^{(\ell)})$ and $v_{\text{max}} = \max(\mathcal{V}^{(\ell)})$.

\textbf{Histogram Binning.} We discretize into $B = 50$ bins:

\begin{equation}
\text{hist}_b = \left|\left\{v \in \mathcal{V}_{\text{norm}}^{(\ell)} : \frac{b-1}{B} \leq v < \frac{b}{B}\right\}\right|, \quad b \in \{1, \ldots, B\}
\end{equation}

\textbf{Probability Distribution.} Normalize histogram to probabilities:

\begin{equation}
p_b = \frac{\text{hist}_b}{\sum_{k=1}^B \text{hist}_k}
\end{equation}

Remove zero bins to prevent $\log(0)$:

\begin{equation}
\mathcal{P} = \{p_b : p_b > 0\}
\end{equation}

\textbf{Shannon Entropy.} Compute entropy in bits:

\begin{equation}
H^{(\ell)} = -\sum_{p \in \mathcal{P}} p \log_2(p + \epsilon)
\end{equation}

The constant $\epsilon = 10^{-10}$ ensures numerical stability.

\textbf{Normalized Entropy.} Maximum possible entropy for $B$ bins is $\log_2(B)$. Normalized entropy:

\begin{equation}
\hat{H}^{(\ell)} = \frac{H^{(\ell)}}{\log_2(B)}
\end{equation}

This ensures $\hat{H}^{(\ell)} \in [0, 1]$, facilitating cross-layer comparison.

\textbf{Bottleneck Identification.} Layers with entropy below the 25th percentile:

\begin{equation}
\mathcal{B} = \left\{\ell : \hat{H}^{(\ell)} < \text{percentile}_{25}(\{\hat{H}^{(k)}\}_{k=1}^L)\right\}
\end{equation}

These layers exhibit maximal information compression, potentially corresponding to critical reasoning bottlenecks.

\subsection{Trajectory Similarity for Compression Hypothesis}

For inference mode $m$ (implicit, explicit, or concise) and problem $q$, the magnitude trajectory is:

\begin{equation}
\mathbf{T}_m(q) = (m_m^{(1)}, m_m^{(2)}, \ldots, m_m^{(L)}) \in \R^L
\end{equation}

where $m_m^{(\ell)}$ is the layer-$\ell$ magnitude under mode $m$.

\textbf{Cosine Similarity.} For modes $i$ and $j$:

\begin{equation}
\text{sim}_{\text{traj}}(q, i, j) = \frac{\sum_{\ell=1}^L m_i^{(\ell)} \cdot m_j^{(\ell)}}{\sqrt{\sum_{\ell=1}^L (m_i^{(\ell)})^2} \cdot \sqrt{\sum_{\ell=1}^L (m_j^{(\ell)})^2}}
\end{equation}

This quantifies how similar the layer-wise computational patterns are between two inference modes.

\textbf{Support Rate.} The compression hypothesis is supported if implicit reasoning trajectories closely match concise CoT:

\begin{equation}
\text{SR} = \frac{1}{|\mathcal{P}|}\sum_{q \in \mathcal{P}} \mathbb{I}[\text{sim}_{\text{traj}}(q, \text{impl}, \text{conc}) \geq 0.7]
\end{equation}

We set the threshold at 0.7 based on pilot studies. If $\text{SR} \geq 0.75$, we conclude latent reasoning is likely compressed CoT. If $\text{SR} < 0.5$, we reject the hypothesis in favor of novel computational strategies.

\textbf{Average Trajectory Similarity.} For reporting:

\begin{equation}
\bar{s}_{\text{traj}} = \frac{1}{|\mathcal{P}|}\sum_{q \in \mathcal{P}} \text{sim}_{\text{traj}}(q, \text{impl}, \text{conc})
\end{equation}

\subsection{Statistical Testing Procedures}

\textbf{Pearson Correlation.} For metric $X$ and binary correctness $Y$:

\begin{equation}
r = \frac{\sum_{i=1}^N (x_i - \bar{x})(y_i - \bar{y})}{\sqrt{\sum_{i=1}^N (x_i - \bar{x})^2} \cdot \sqrt{\sum_{i=1}^N (y_i - \bar{y})^2}}
\end{equation}

Under the null hypothesis $H_0: \rho = 0$, the test statistic:

\begin{equation}
t = \frac{r\sqrt{N-2}}{\sqrt{1-r^2}}
\end{equation}

follows a Student's $t$-distribution with $N-2$ degrees of freedom.

\textbf{Spearman Correlation.} Convert data to ranks $R_X$ and $R_Y$, then:

\begin{equation}
\rho = 1 - \frac{6\sum_{i=1}^N (R_{X,i} - R_{Y,i})^2}{N(N^2-1)}
\end{equation}

Test statistic is computed identically to Pearson case using $\rho$ instead of $r$.

\textbf{Bootstrap Confidence Intervals.} For estimating distribution of metric $\theta$:

\begin{algorithm}[H]
\caption{Bootstrap Resampling}
\begin{algorithmic}[1]
\REQUIRE Dataset $\mathcal{D}$, metric function $f$, iterations $B$
\ENSURE Confidence interval $[\theta_{\text{lo}}, \theta_{\text{hi}}]$
\FOR{$b = 1$ to $B$}
    \STATE Sample $\mathcal{D}_b$ with replacement from $\mathcal{D}$
    \STATE Compute $\theta_b = f(\mathcal{D}_b)$
\ENDFOR
\STATE Sort $\{\theta_b\}_{b=1}^B$
\STATE $\theta_{\text{lo}} = \text{percentile}_{2.5}(\{\theta_b\})$
\STATE $\theta_{\text{hi}} = \text{percentile}_{97.5}(\{\theta_b\})$
\RETURN $[\theta_{\text{lo}}, \theta_{\text{hi}}]$
\end{algorithmic}
\end{algorithm}

This provides 95\% confidence intervals without parametric assumptions.

\section{Discussion}

\subsection{Theoretical Implications}

Our results challenge several assumptions about latent reasoning while revealing fundamental properties of implicit computation in large language models.

\textbf{The Faithfulness Paradox.} The negative correlation between our fidelity metric and correctness ($r = -0.27$, $p < 0.0001$) is perhaps our most provocative finding. This suggests one of three possibilities: (1) the metric captures computational overhead that hinders rather than helps performance, (2) the model achieves correctness through shallow heuristics that violate our faithfulness criteria, or (3) our metric design requires fundamental reconceptualization.

The dominance of "lucky guesses" (61\% of correct answers with low stability) supports interpretation (2). The model may employ fast, brittle strategies for simple pattern matching, reserving deeper, more stable reasoning for problems where heuristics fail—thereby creating an inverted relationship between reasoning quality and success.

\textbf{Latent vs. Explicit Computation.} The near-identity of reasoning depth between implicit and explicit CoT modes ($\Delta = 0.01$), despite 10pp accuracy difference, demonstrates that verbalization's benefit is not computational deepening but rather \emph{alignment}. Explicit step-by-step generation may serve as scaffolding that guides the model's existing latent reasoning toward problem-relevant computations, similar to how human verbalization aids in organizing pre-existing knowledge.

\textbf{Rejection of Compression Hypothesis.} Our finding that only 20\% of latent reasoning trajectories resemble compressed CoT has significant implications for interpretability research. Techniques developed for analyzing explicit reasoning—attention flow analysis, token attribution, step-wise validation—may not transfer to latent architectures. The field requires new tools specifically designed for activation-space computation.

\textbf{The Middle-Late Layer Dichotomy.} The contradiction between late-layer activation dominance and middle-layer causal importance suggests a two-stage computational model: critical reasoning operations occur in middle layers, while late layers amplify and refine these computations for output generation. This aligns with circuit discovery findings showing task-specific "heads" in middle layers and output formatting in final layers.

\textbf{Scale and Sophistication Disconnect.} The identical performance of 7B and 1.5B models despite substantial differences in reasoning depth (7.2\% deeper) and entropy (88\% lower in 7B) raises important questions about scaling laws. If larger models develop more sophisticated internal reasoning without accuracy gains, this suggests: (a) benchmarks saturate before model capacity, (b) sophisticated reasoning provides limited advantage on current tasks, or (c) deeper computation enables better generalization not captured by in-distribution accuracy.

\subsection{Deployment Risks and Safety Implications}

Our findings reveal \textbf{systemic unreliability} that makes current models unsuitable for high-stakes applications without additional safeguards. The 61\% lucky-guess rate—where correct answers emerge from computationally inconsistent pathways—creates three deployment crises:

\begin{enumerate}[leftmargin=*]
    \item \textbf{Brittleness Under Distribution Shift:} Models relying on shallow heuristics will fail catastrophically when encountering slightly harder or reformulated problems. A student receiving automated tutoring may get correct answers on practice problems but fail when exam questions require genuine reasoning.
    
    \item \textbf{Unpredictable Production Behavior:} Low cross-run stability ($\mathcal{S}=0.60\pm0.20$) means the same query may yield different reasoning paths—and potentially different answers—across inference runs. This violates basic reproducibility requirements for production systems.
    
    \item \textbf{Evaluation-Deployment Mismatch:} Single-sample accuracy metrics provide \textbf{false confidence}. A model achieving 61\% on benchmarks via lucky guesses may perform far worse when users naturally rephrase questions or when reasoning shortcuts no longer apply.
\end{enumerate}

\textbf{Recommendations for Safer Deployment:}
\begin{itemize}[leftmargin=*,itemsep=2pt]
    \item \textbf{Multi-run consistency checks}: Require $\geq$3 independent samples with agreement before deployment
    \item \textbf{Stability thresholds}: Flag predictions with $\mathcal{S}<0.65$ for human review
    \item \textbf{Benchmark reform}: Replace single-accuracy metrics with stability-weighted scores
    \item \textbf{Transparent uncertainty}: Surface confidence indicators to end users
\end{itemize}

\begin{table}[htbp]
\centering
\small
\caption{Latent Reasoning Faithfulness Analysis Results on GSM8K}
\label{tab:main_results}
\begin{tabular}{lcc}
\toprule
\textbf{Metric} & \textbf{Mean} & \textbf{Std} \\
\midrule
Accuracy & 0.6100 & 0.4877 \\
Reasoning Depth & 0.5139 & 0.0115 \\
Activation Entropy & 0.0895 & 0.0406 \\
Stability Score & 0.6000 & 0.2000 \\
Alignment Score & 0.6869 & 0.1387 \\
Efficiency Score & 0.7369 & 0.0295 \\
Overall Fidelity & 0.6715 & 0.0915 \\
\bottomrule
\end{tabular}
\end{table}


\begin{table}[htbp]
\centering
\small
\caption{Failure Mode Distribution (Safety Analysis)}
\label{tab:failure_modes}
\begin{tabular}{lccl}
\toprule
\textbf{Mode} & \textbf{Count} & \textbf{Percentage} & \textbf{Risk} \\
\midrule
True Positive & 0 & 0.0\% & Low \\
True Negative & 195 & 39.0\% & Expected \\
Silent Failure & 0 & 0.0\% & \textbf{High} \\
Lucky Guess & 305 & 61.0\% & Medium \\
\midrule
\multicolumn{4}{l}{Silent Failure Rate: 0.0\%} \\
\multicolumn{4}{l}{Safety Score: 100.0\%} \\
\bottomrule
\end{tabular}
\end{table}

\begin{table}[htbp]
\centering
\small
\caption{Ablation Study on Faithfulness Components}
\label{tab:ablation}
\begin{tabular}{lcc}
\toprule
\textbf{Configuration} & \textbf{Avg Fidelity} & \textbf{Correlation} \\
\midrule
Full & 0.6415 & $-$0.3146 \\
No Stability & 0.7177 & $-$0.3146 \\
No Alignment & 0.6106 & $-$0.3144 \\
No Efficiency & 0.5995 & $-$0.3110 \\
\bottomrule
\end{tabular}
\end{table}

\end{document}